\documentclass{article}
\usepackage{arxiv}

\usepackage[utf8]{inputenc} 
\usepackage[T1]{fontenc}    
\usepackage{hyperref}       
\usepackage{url}            
\usepackage{wrapfig}
\usepackage{booktabs}       
\usepackage{amsfonts}       
\usepackage{nicefrac}       
\usepackage{microtype}      
\usepackage{lipsum}		
\usepackage{tikz}
\usetikzlibrary{arrows.meta, positioning, fit, calc, backgrounds}
\usepackage{standalone}
\usepackage{graphicx}
\usepackage{natbib}
\usepackage{doi}

\usepackage{lmodern}
\usepackage{amsmath, amssymb, amsthm}
\usepackage{mathtools}
\usepackage{booktabs}
\usepackage{multirow}
\usepackage{array}
\usepackage{caption}
\usepackage{subcaption}
\usepackage{xcolor}
\usepackage{enumitem}
\usepackage{algorithm2e}
\usepackage{times}
\usepackage{svg}
\usepackage{makecell}
\usepackage{hyperref}

\usepackage{listings}
\hypersetup{
  colorlinks=true,
  linkcolor=blue,
  citecolor=blue,
  urlcolor=blue,
  pdftitle={Attention Editing: Progressive Distillation for Cross-Architecture Attention Replacement},
  pdfauthor={Anonymous Authors}
}

\title{Attention Editing: A Versatile Framework for Cross-Architecture Attention Conversion}

\author{
Zhen Cheng \quad Hao-Bo Yang \quad Wan-Yi Huang \quad Jin-Long Li \\
China Merchants Bank Artificial Intelligence Laboratory \\
\texttt{\{chengzhen1005, yanghaobo, huang\_wanyi, lucida\}@cmbchina.com}
}

\renewcommand{\headeright}{}
\renewcommand{\undertitle}{Technical Report}

\hypersetup{
pdftitle={A template for the arxiv style},
pdfsubject={q-bio.NC, q-bio.QM},
pdfauthor={David S.~Hippocampus, Elias D.~Striatum},
pdfkeywords={First keyword, Second keyword, More},
}

\begin{document}
\maketitle

\begin{abstract}
Key--Value (KV) cache memory and bandwidth increasingly dominate large language model inference cost in long-context and long-generation regimes. Architectures such as multi-head latent attention (MLA) and hybrid sliding-window attention (SWA) can alleviate this bound, but integrating them into existing models remains difficult. Prior methods impose fine-grained structural requirements on both source and target attention modules, which cannot meet the feasible requirement in practical deployment.
We present \textit{\textbf{Attention Editing}}, a practical framework for converting already-trained large language models (LLMs) with new attention architectures without re-pretraining from scratch. Attention editing replaces the original attention with a learnable target module and trains it using \textit{\textbf{progressive distillation}}, consisting of (1) layer-wise teacher-forced optimization with intermediate activation supervision to prevent cold-start error accumulation, and (2) model-level distillation on next-token distributions, optionally regularized by weak feature matching. We instantiate the framework on \textit{two different targets—MLA and GateSWA}, a gated hybrid SWA design, and apply it to Qwen3-8B and Qwen3-30B-A3B. The resulting models maintain competitive performance while delivering substantial efficiency improvements, demonstrating that large-scale attention conversion is both feasible and robust. Notably, experiments are conducted on an Ascend 910B clusters, offering a practical training case study on domestic hardware.
\end{abstract}

\section{Introduction}

Agentic large language model (LLM) applications, from tool-using assistants to end-to-end AGI agents, increasingly execute \emph{long interactive trajectories} that interleave user instructions, intermediate reasoning, tool calls, and tool observations \citep{yao2022react, schick2023toolformer,openai2025gptoss_modelcard}. Such workflows naturally push both the \emph{input context} (conversation history, retrieved evidence, and tool outputs) and the \emph{output length} (multi-step plans, long-form explanations, and iterative self-correction) upward. Meanwhile, recent open-weight reasoning and agent-capable models explicitly target these use cases and emphasize controllable ``reasoning effort'' or ``thinking'' behavior \citep{openai2025gptoss_modelcard, qwen2025qwen3}. In this regime, the inference bottleneck is increasingly memory- and bandwidth-bound: autoregressive decoding caches per-layer keys and values for all previous tokens, and the KV cache grows with sequence length and batch size \citep{kwon2023pagedattention}.

To alleviate KV-cache overhead, the community has developed a wide spectrum of efficient attention mechanisms. At the architecture level, \textit{Multi-head Latent Attention (MLA)} compresses key/value states into a low-rank latent representation, substantially reducing cached memory while retaining expressiveness \citep{liu2024deepseekv2,deepseekv3,deepseekr1,kimiteam2025kimik2openagentic,glm5team2026glm5}. \textit{Linear attention} and its variants redesign attention to avoid quadratic scaling and can change the structure of state carried across long contexts \citep{katharopoulos2020linear, choromanski2021performer,peng2023rwkv,mamba,yang2025gateddeltanet}. In parallel, \textit{sliding-window attention (SWA)} restrict attention to a local window for most layers (often with periodic global layers), achieving favorable compute and cache scaling for long sequences \citep{beltagy2020longformer, openai2025gptoss_modelcard, xiao2026mimov2flash,stepfun2026step35flash}. However, adopting a new architecture typically requires training from scratch. This has motivated post-hoc conversion methods such as TransMLA and MHA2MLA that aim to obtain MLA's KV-cache benefits while reusing existing model weights \citep{meng2025transmla, ji2025mha2mla}.

However, existing approaches face three practical limitations for real-world applications. \textbf{First}, many conversions impose \emph{fine-grained structural requirements} on both source and target attention modules, because they rely on matrix factorization / low-rank linear approximations (e.g., SVD-based projections) and attention-specific handling of positional encoding (e.g., RoPE variants) to initialize or constrain transformed weights \citep{ji2025mha2mla, bercovich2025puzzle,koike2026latentllm}. In practice, deployed model implementations are often flexible (e.g., mixing kernel variants, hybrid attention patterns, MoE blocks, and serving-optimized layouts), which can deviate from the assumptions required by a given conversion recipe. \textbf{Second}, conversions are frequently demonstrated on \emph{base model} checkpoints and require re-running substantial post-training (SFT/RL) to recover chat/reasoning behavior, potentially discarding expensive gains from post-training stage \citep{ouyang2022instructgpt, qwen2025qwen3, meng2025transmla}. \textbf{Third}, successful recovery can heavily depend on training data quality and distribution: Both TransMLA and MHA2MLA adopt the SmolLM series as their primary backbone models, owing to the availability of its open-source pretraining corpus. Since modern LLMs are pretrained on proprietary mixtures, practitioners often cannot obtain original pre-training corpus, making data-efficient and distribution-robust editing essential.

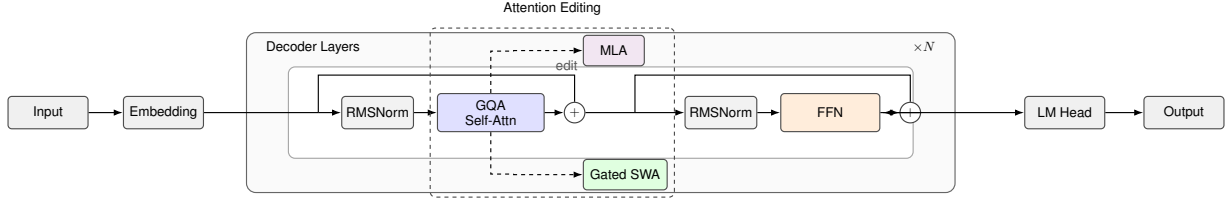
\begin{figure}[t]
    \centering
    \begin{tikzpicture}[
    font=\sffamily,
    every node/.style={font=\sffamily},
    >={Latex[length=2.4mm,width=1.7mm]},
    line width=0.85pt,
    node distance=6mm and 8mm,
    box/.style={
        draw=black!70,
        rounded corners=2.8pt,
        minimum height=8.5mm,
        align=center
    },
    io/.style={box, fill=gray!12, minimum width=21mm},
    norm/.style={box, fill=gray!10, minimum width=18mm},
    attn/.style={box, fill=blue!12, minimum width=28mm, minimum height=10mm},
    ffn/.style={box, fill=orange!14, minimum width=26mm, minimum height=10mm},
    altA/.style={box, fill=violet!10, minimum width=16mm, minimum height=8mm},
    altB/.style={box, fill=green!12, minimum width=22mm, minimum height=8mm},
    plus/.style={
        circle,
        draw=black!75,
        fill=white,
        inner sep=0pt,
        minimum size=5.4mm
    },
    stackbg/.style={
        draw=black!60,
        rounded corners=7pt,
        fill=gray!4,
        inner sep=8pt
    },
    layerbg/.style={
        draw=black!35,
        rounded corners=5pt,
        fill=white
    },
    editbox/.style={
        draw=black!70,
        dashed,
        rounded corners=4pt,
        inner sep=5pt
    }
]

\node[io] (input) {Input};
\node[io, right=9mm of input] (embed) {Embedding};

\node[stackbg, right=11mm of embed, minimum width=186mm, minimum height=42mm] (stack) {};
\node[anchor=west] at ($(stack.north west)+(4mm,-4mm)$) {Decoder Layers};
\node[anchor=east] at ($(stack.north east)+(-4mm,-4mm)$) {$\times N$};

\node[layerbg, minimum width=164mm, minimum height=24mm] (layer) at (stack.center) {};

\node[io, right=18mm of stack] (head) {LM Head};
\node[io, right=10mm of head] (output) {Output};

\coordinate (xin)  at ($(layer.west)+(8mm,0)$);
\coordinate (xout) at ($(layer.east)+(-8mm,0)$);

\node[norm, anchor=west] (norm1) at ($(layer.west)+(14mm,0)$) {RMSNorm};
\node[attn, right=6mm of norm1] (attn) {GQA\\Self-Attn};
\node[plus, right=5mm of attn] (add1) {$+$};

\node[norm, right=26mm of add1] (norm2) {RMSNorm};

\node[ffn, right=6mm of norm2] (ffn) {FFN};
\node[plus, right=5mm of ffn] (add2) {$+$};

\draw[->] (input.east) -- (embed.west);
\draw[-] (embed.east) -- (xin);
\draw[->] (xin) -- (norm1.west);
\draw[->] (norm1.east) -- (attn.west);
\draw[->] (attn.east) -- (add1.west);

\coordinate (xmid) at ($(add1.east)!0.5!(norm2.west)$);
\draw[-] (add1.east) -- (xmid);
\draw[->] (xmid) -- (norm2.west);

\draw[->] (norm2.east) -- (ffn.west);
\draw[->] (ffn.east) -- (add2.west);
\draw[->] (add2.east) -- (xout);
\draw[->] (xout) -- (head.west);
\draw[->] (head.east) -- (output.west);

\draw (xin)  |- ($(add1.north)+(0,7mm)$) -| (add1.north);
\draw (xmid) |- ($(add2.north)+(0,7mm)$) -| (add2.north);

\node[altA, above right=7mm and 10mm of attn] (mla) {MLA};
\node[altB, below right=7mm and 10mm of attn] (gswa) {Gated SWA};

\coordinate (editup)     at ($(attn.south)-(0,11mm)$);
\coordinate (editbranch) at ($(editup)+(6mm,0)$);
\coordinate (editdown)     at ($(attn.north)+(0,11mm)$);
\coordinate (editbranch2) at ($(editdown)+(6mm,0)$);

\draw[dashed] (attn.south) -- (editup) -- (editbranch);
\draw[dashed] (attn.north) -- (editdown) -- (editbranch2);
\draw[->, dashed] (editbranch2) |- (mla.west);
\draw[->, dashed] (editbranch) |- (gswa.west);

\node[text=black!65] at ($(attn.east)!0.58!(mla.west)+(0,3mm)$) {edit};

\node[editbox, fit=(attn)(mla)(gswa)] (editfit) {};
\node[above=2mm of editfit] {Attention Editing};

\end{tikzpicture}
    \caption{Illustration of attention editing. Attention editing is a general framework for substantially modifying the attention architecture of an \emph{already-trained} LLM without re-pretraining from scratch. It does not depend on delicate weight surgery, and treats the target attention modules as \emph{learnable replacements}.}
    \label{fig:attention_editing}
    \vspace{-0.10in}
\end{figure}

To address these limitations, we introduce \textbf{Attention Editing}, shown in Figure \ref{fig:attention_editing}, a general framework for substantially converting the attention architecture of an already-trained LLM without training from scratch. Unlike prior recipes that depend on delicate weight surgery, we treat the target attention modules as \emph{learnable replacements}: we allow most parameters inside the new attention modules to be \emph{randomly initialized} and then trained efficiently via \textbf{progressive distillation}. Concretely, we propose a two-stage recipe. (i) \textbf{Layer-wise teacher forcing} trains each decoder layer using teacher-provided intermediate activations to avoid deep error accumulation at cold start, bringing the newly initialized attention weights to a strong working point (intermediate ``hints'' style supervision) \citep{romero2014fitnets}. (ii) We then apply \textbf{model-level distillation} by matching teacher and student next-token distributions with KL divergence (knowledge distillation) \citep{hinton2015distill}. We optionally add weak intermediate-feature matching as regularization.


We validate Attention Editing by converting GQA$\rightarrow$MLA on \texttt{Qwen3-8B} and \texttt{Qwen3-30B-A3B} \citep{qwen2025qwen3}. To demonstrate generality beyond MLA, we also edit GQA into \textbf{Gate Sliding-Window Attention (GateSWA)}: a tiny-window sliding window attention (SWA) hybrid inspired by combining GPT-OSS and Qwen3-Next \citep{openai2025gptoss_modelcard,qwen3next80ba3b}. Motivated by recent evidence that a simple gate applied to the attention output can mitigate attention sinks and improve stability, we remove explicit learnable sink biases/tokens and instead adopt an \emph{element-wise gate} on the SWA output\citep{xiao2023streamingllm,qiu2025gatedattention}. We apply Attention Editing to convert GQA-based models into GateSWA-GQA using a 5:1 sliding-to-full schedule. Results indicate competitive quality with strong hardware-efficiency gains. All training runs are conducted on an Ascend 910B cluster, and our setup follows the growing evidence that large-scale model training on Ascend clusters is feasible in practice.

These two targets (MLA and GateSWA) jointly support our claims that attention can be substantially refactored post-training and that progressive distillation enables data-efficient learning from random-init modules. Our contributions can be summarized as follows:
\begin{itemize}
  \item We formalize \textbf{Attention Editing}: fundamental attention-architecture changes are feasible for already-trained LLMs, without the requirement of delicate weight surgery.
  \item We propose \textbf{progressive distillation} as a robust, structure-insensitive, and data-efficient method for attention editing.
  \item We introduce \textbf{GateSWA}, an efficient hybrid attention variant that replaces learnable sink mechanisms in hybrid SWA models with an element-wise gate.
  \item We present a practical case study of large-model attention editing trained entirely on Ascend 910B, providing an actionable recipe for large-scale training on domestic hardware.
\end{itemize}

\section{Related Works}
\label{sec:related}

\paragraph{Efficient attention architectures.}
To reduce the KV-cache memory occupied by MHA, MQA \citep{shazeer2019fast} and GQA \citep{ainslie2023gqa} are designed to share KV heads to reduce decoding bandwidth. Furthermore, a large body of work studies the trade-off between model quality and inference efficiency by redesigning the attention mechanism, especially to reduce KV-cache memory and bandwidth during autoregressive decoding. Representative directions include \textit{linear attention}, which replaces softmax attention with kernelized or recurrent formulations to improve scaling with sequence length \citep{katharopoulos2020linear, choromanski2021performer,peng2023rwkv,mamba,yang2025gateddeltanet,zhang2025kimilinear}; \textit{Multi-head Latent Attention}, which compresses key/value states into a low-rank latent representation and has been adopted in recent reasoning-oriented model families \citep{liu2024deepseekv2, deepseekv3, deepseekr1}; and \textit{sliding-window or hybrid attention}, which restricts most layers to local attention while retaining periodic global layers for long-context modeling \citep{beltagy2020longformer, openai2025gptoss_modelcard, xiao2026mimov2flash}. These approaches are complementary to other efficiency techniques like KV-cache quantization or system-level serving optimizations, which are important but orthogonal to the attention-architecture focus of this work.

\paragraph{Attention architecture conversion.}
Since replacing the attention module usually requires expensive re-training, recent work has explored post-hoc conversion from one attention architecture form to another. In particular, TransMLA and MHA2MLA study conversion from GQA/MHA to MLA in order to inherit MLA's KV-cache advantages while reusing existing model weights \citep{meng2025transmla, ji2025mha2mla}. Besides that, the other work from NVIDIA Nemotron treat attention configuration itself, reducing KV heads or modifying GQA layouts, as an important efficiency knob in deployment-oriented model design \citep{bercovich2025puzzle,bercovich2025llamanemotron}. Compared with these methods, our goal is different: rather than relying on attention-specific weight surgery or fine-grained structural assumptions, we view the target attention as a learnable replacement, enabling much greater architectural changes and making the approach applicable not only to base models but also to already post-trained chat or reasoning LLMs.

\paragraph{Knowledge distillation.}
Knowledge distillation transfers behavior by matching teacher and student output distributions \citep{hinton2015distill}, while intermediate-feature or hint-based distillation supervises hidden representations to stabilize training when the student architecture differs substantially from the teacher \citep{romero2014fitnets}. For autoregressive language models, recent work further emphasizes \emph{on-policy} or generation-aware distillation to reduce the mismatch between teacher-forced training and student-generated inference trajectories \citep{agarwal2024gkd, gu2023minillm}. Our method combines these lines into a progressive recipe: hidden-state distillation first brings randomly initialized replacement modules to a workable regime, and output-level distillation then refines next-token behavior. 

\section{Preliminary}
\label{sec:problem}

\subsection{Multi Head Softmax Attention}

Let $h_t \in \mathbb{R}^{d}$ denote the input representation at position $t$, $n_h$ the number of attention heads, and $d_h$ the per-head dimension. In standard multi-head attention (MHA) \citep{vaswani2017attention}, we first compute
\begin{align}
q_t &= W^{Q} h_t, \\
k_t &= W^{K} h_t, \\
v_t &= W^{V} h_t,
\end{align}
where $q_t,k_t,v_t \in \mathbb{R}^{n_h d_h}$. We then split them into $n_h$ heads, i.e.,
\begin{align}
[q_{t,1}; q_{t,2}; \cdots; q_{t,n_h}] &= q_t, \\
[k_{t,1}; k_{t,2}; \cdots; k_{t,n_h}] &= k_t, \\
[v_{t,1}; v_{t,2}; \cdots; v_{t,n_h}] &= v_t,
\end{align}
with $q_{t,i}, k_{t,i}, v_{t,i} \in \mathbb{R}^{d_h}$. For causal self-attention, the output of the $i$-th head at step $t$ is
\begin{align}
\alpha_{t,i,j}
&=
\frac{\exp\!\left(q_{t,i}^{\top} k_{j,i} / \sqrt{d_h}\right)}
{\sum_{s=1}^{t} \exp\!\left(q_{t,i}^{\top} k_{s,i} / \sqrt{d_h}\right)},
\qquad 1 \le j \le t, \\
o_{t,i}
&=
\sum_{j=1}^{t} \alpha_{t,i,j} v_{j,i},
\end{align}
and the final attention output is obtained by concatenating all heads and applying the output projection,
\begin{align}
u_t = W^{O}[o_{t,1}; o_{t,2}; \cdots; o_{t,n_h}].
\end{align}
Equivalently, if we stack all positions into matrices $Q_i$, $K_i$, and $V_i$ for the $i$-th head, then $O_i=\mathrm{softmax}(Q_i K_i^{\top}/\sqrt{d_h})V_i$, which matches the standard scaled dot-product attention form.

\subsection{Efficient Attention Architectures}

To reduce the decoding-time memory cost of KV-cache, several efficient attention variants modify the above computation while retaining the same head-level notation. We next review three efficient attention variants.

\paragraph{Multi-Head Latent Attention (MLA).} Instead of caching all per-token keys and values, MLA compresses them into a shared latent representation:
\begin{align}
c_t^{KV} &= W^{DKV} h_t, \\
k_t^{C} &= W^{UK} c_t^{KV}, \\
v_t^{C} &= W^{UV} c_t^{KV},
\end{align}
where $c_t^{KV} \in \mathbb{R}^{d_c}$ is a low-dimensional latent code with $d_c \ll n_h d_h$. To remain compatible with rotary position embeddings, MLA further introduces decoupled positional components $q_t^{R}$ and $k_t^{R}$, and computes attention with concatenated queries and keys, e.g., $q_{t,i}=[q_{t,i}^{C}; q_{t,i}^{R}]$ and $k_{j,i}=[k_{j,i}^{C}; k_{j}^{R}]$. The key idea is to replace KV-cache of each token in the original head space by a shared low-rank latent cache, so that the per-token cache over $\ell$ layers becomes $(d_c + d_h^{R})\ell$ rather than $2 n_h d_h \ell$.

\paragraph{Linear hybrid attention.} Linear attention replaces the growing prefix-wise KV cache with a fixed-size recurrent memory. For clarity, we write the update rule for a single head. A broad class of linear attention methods maintains a finite-state memory $S_t$ instead of the full prefix KV cache:
\begin{align}
S_t &= A_t S_{t-1} + k_t v_t^{\top}, \\
o_t &= S_t^{\top} q_t,
\end{align}
where $A_t$ is a data-dependent or data-independent transition matrix that controls forgetting and gating. Different linear attention variants mainly differ in how $A_t$ is parameterized.
In a hybrid architecture, linear-attention layers are interleaved with a subset of full-attention layers. The linear layers only maintain fixed-size recurrent states, whose memory cost does not grow with the context length, while only the full-attention layers keep token-wise KV caches. For example, Kimi Linear \citep{zhang2025kimilinear} instantiates this design by interleaving Kimi Delta Attention (KDA) and MLA, and delivers substantial reductions in KV-cache memory relative to full attention.

\paragraph{Sliding-window attention (SWA).} SWA restricts each query to a local window of size $w$, namely
\begin{align}
o_{t,i}^{\mathrm{SWA}}
=
\sum_{j=\max(1,\, t-w+1)}^{t}
\alpha_{t,i,j} v_{j,i},
\end{align}
where the softmax normalization is performed only over the visible window. After the cache is filled, SWA requires only a rolling buffer of size $O(2 w n_h d_h)$ per layer, instead of a cache that grows linearly with the full prefix length. A practical challenge in streaming SWA is the \emph{attention sink}, namely that certain early positions absorb disproportionate attention mass even when they carry little semantic information \citep{xiao2023streamingllm,gu2025attentionsink}. To tackle the influence of attention sink, one remedy is to introduce a dedicated learnable sink token during training, so that excess attention mass is redirected to a parameterized placeholder rather than to ordinary context tokens. Another remedy is gated attention, which applies a query-dependent gate to the SDPA output, e.g.,
\begin{align}
\tilde{o}_{t,i} = g_{t,i} \odot o_{t,i},
\qquad
g_{t,i} = \sigma(W_g h_t),
\end{align}
thereby suppressing query-irrelevant attention outputs. Recent evidence shows that such post-SDPA gating can mitigate attention sink and improve long-context extrapolation \citep{qiu2025gatedattention}.

In the experimental setup of this work, we ultimately select GateSWA and MLA as the target architectures. On the one hand, these attentions have relatively stable inference implementations in the community. On the other hand, they are better aligned with the Ascend clusters we employ, thereby facilitating more efficient experimentation.

\section{Method: Progressive Distillation for Attention Editing}
\label{sec:method}

We consider \emph{attention editing}, where the chat/reasoning LLM is converted into a new architecture by replacing the attention architecture, while keeping the remainder of the network as intact as possible. Let $\mathcal{T}$ denote the original pretrained model (the \emph{teacher}) and $\mathcal{S}$ the edited model (the \emph{student}). We write the student parameters as
\begin{align}
\theta^{\mathcal{S}} = \theta_{\mathrm{keep}} \cup \theta_{\mathrm{edit}},
\end{align}
where $\theta_{\mathrm{keep}}$ contains parameters that remain structurally compatible with the teacher and are copied from $\mathcal{T}$, and $\theta_{\mathrm{edit}}$ contains parameters introduced by the edited attention modules. We intentionally initialize $\theta_{\mathrm{edit}}$ at random, rather than deriving them from a carefully structured matrix decomposition of pretrained attention weights.

\begin{figure}[t]
    \centering
    \includegraphics[width=0.6\linewidth]{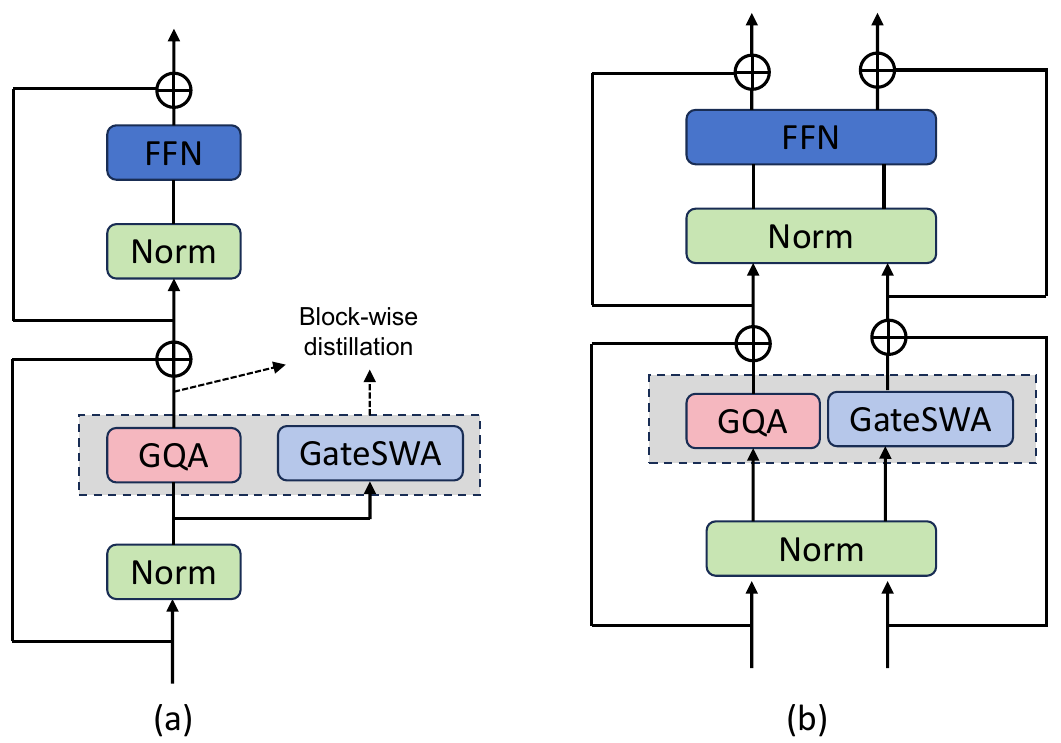}
    \caption{Illustration of the forward propogation in two stages of progressive distillation. (a) Block-wise teacher forcing distillation: the inputs to each layer comes from the original model; (b) Model-level distillation: each architecture maintain its own inputs. \textit{The gray shading indicates that these two modules can be merged into a single model and trained efficiently using sharding strategies such as FSDP or DeepSpeed.}}
    \label{fig:method_illustration}
    \vspace{-0.10in}
\end{figure}

This design choice distinguishes our setting from prior conversion methods such as TransMLA and MHA2MLA, which exploit strong structural constraints and explicitly construct initial MLA parameters from the original attention weights via low-rank factorization or related decompositions~\citep{meng2025transmla, ji2025mha2mla}. While such initialization is elegant and effective when the source and target attention forms are tightly aligned, it becomes increasingly restrictive when the architectural edit is large. Our goal is therefore different: we seek an optimization procedure that remains effective even when the edited attention differs substantially from the original one. Randomly initializing the edited attention parameters avoids imposing a brittle source-to-target correspondence and allows the new attention modules to be optimized more freely. At the same time, we still preserve structurally compatible parameters whenever possible. The exact retained weights is described in Section~\ref{sec:equipment_mla_gateswa}.

\subsection{Problem Setup}

Let $\mathcal{D}=\{x^{(i)}\}_{i=1}^{N}$ denote the training corpus, where each sequence
\begin{align}
x^{(i)} = (x^{(i)}_1,\dots,x^{(i)}_{n_i})
\end{align}
is presented in standard pre-training format. Unlike instruction-tuning distillation, our data do not contain an explicit prompt/answer split, and we therefore apply supervision to \emph{all} valid next-token positions instead of restricting the loss to response spans only.

Let $L$ be the number of Transformer blocks, and let $\mathcal{B}\subseteq\{1,\dots,L\}$ denote the set of layers whose attention modules are edited. For a token position $t$ and layer $\ell$, $h_{t}^{\mathcal{T},\ell}$ and $h_{t}^{\mathcal{S},\ell}$ are denoted as the hidden states of the teacher and student, respectively. We further denote by
\begin{align}
u_{t}^{\mathcal{T},\ell},\quad u_{t}^{\mathcal{S},\ell}\in\mathbb{R}^{d}
\end{align}
the output of the attention branch \emph{after} the output projection (\texttt{o\_proj}) and \emph{before} residual addition at layer $\ell$. This quantity is the basic intermediate representation used in our first training stage.

A natural way to recover the chat and reasoning ability of the teacher is to distill the student from the teacher by minimizing a token-level divergence between their output distributions, following standard knowledge distillation~\citep{hinton2015distill}. However, in our setting, directly applying model-level distillation from the beginning is highly unstable. Since the edited attention parameters in shallow layers are randomly initialized, the student receives poor fine-grained signals, and the resulting representation error accumulates rapidly with depth. In practice, this causes the optimization of output-level distillation to stall. To address this issue, we propose \emph{progressive distillation}: we first align intermediate representations under a block-wise teacher-forcing regime to obtain a reliable initialization, and only then transition to end-to-end distillation on the model outputs.

\subsection{Progressive Distillation}

Our training procedure consists of two stages:
\begin{enumerate}
    \item \textbf{Block-wise teacher-forcing distillation.} Each edited attention block is trained independently using the teacher's hidden state as input, so that early errors from an under-trained student block do not propagate to deeper layers.
    \item \textbf{Model-level distillation.} After the edited blocks acquire a meaningful initialization, the entire student is trained end-to-end using token-level knowledge distillation, optionally augmented with a small intermediate-state similarity loss.
\end{enumerate}

The key idea is to move from \emph{local} matching to \emph{global} matching. Stage I turns the difficult problem of optimizing a randomly initialized edited network into a collection of well-conditioned local regression problems. Stage II then restores full autoregressive behavior by matching the teacher's output distribution under the student’s own forward dynamics, as shown in Figure \ref{fig:method_illustration}.

\paragraph{Stage I: Block-Wise Teacher-Forcing Distillation.}

For each edited layer $\ell\in\mathcal{B}$, we construct a block-wise training problem in which the input to the edited student attention block is clamped to the teacher hidden state. Concretely, let
\begin{align}
H^{\mathcal{T},\ell-1} = [h_1^{\mathcal{T},\ell-1}; \dots; h_n^{\mathcal{T},\ell-1}] \in \mathbb{R}^{n\times d}
\end{align}
be the teacher representation entering layer $\ell$. We then feed $H^{\mathcal{T},\ell-1}$ into the edited attention block of the student and obtain
\begin{align}
\widehat{U}^{\mathcal{S},\ell}
=
\mathcal{A}^{\mathcal{S}}_{\ell}(H^{\mathcal{T},\ell-1}),
\end{align}
where $\mathcal{A}^{\mathcal{S}}_{\ell}$ denotes the edited attention branch up to and including \texttt{o\_proj}. The corresponding target is the teacher attention-branch output
\begin{align}
U^{\mathcal{T},\ell}
=
\mathcal{A}^{\mathcal{T}}_{\ell}(H^{\mathcal{T},\ell-1}).
\end{align}

We supervise the edited block by matching these post-\texttt{o\_proj} activations, using Mean Square Error (MSE) loss. To improve numerical stability, we normalize the MSE loss before computing the regression loss as \citep{bercovich2025puzzle}. The block-wise loss for layer $\ell$ is
\begin{align}
\mathcal{L}_{\mathrm{blk}}^{(\ell)}
=
\frac{
\left\|
\widehat{U}^{\mathcal{S},\ell}
-
U^{\mathcal{T},\ell}
\right\|_F^2
}{
\left\|
U^{\mathcal{T},\ell}
\right\|_F^2 + \epsilon
}.
\label{eq:block_loss}
\end{align}
In practice, we optimize each edited layer independently. That is, when training layer $\ell$, the supervision signal is always computed from the teacher input $H^{\mathcal{T},\ell-1}$ rather than from representations produced by previously edited student layers. This block-wise teacher-forcing strategy removes the main source of optimization failure in direct distillation: poor signals from shallow edited blocks can no longer cascade through the network and corrupt deeper targets.

The choice of the post-\texttt{o\_proj} output as the regression target is deliberate. First, it is the quantity that is directly injected into the residual stream, and therefore most immediately controls the downstream hidden-state trajectory. Second, unlike attention logits or head-wise $QK^\top$ statistics, it remains well defined even when the source and target attention parameterizations differ substantially. This makes it a robust architecture-agnostic intermediate target for attention editing.

\paragraph{Stage II: Model-Level Distillation.}
At this stage, the student is sufficiently well initialized for output-level distillation to become effective. Let $z_t^{\mathcal{T}}$ and $z_t^{\mathcal{S}}$ denote the teacher and student logits at position $t$, respectively. We use the standard token-level distillation objective~\citep{hinton2015distill}
\begin{align}
p_t^{\mathcal{T}}
&=
\mathrm{softmax}\!\left(z_t^{\mathcal{T}}/\tau\right),\\
p_t^{\mathcal{S}}
&=
\mathrm{softmax}\!\left(z_t^{\mathcal{S}}/\tau\right),
\end{align}
and define
\begin{align}
\mathcal{L}_{\mathrm{KD}}
=
\frac{\tau^2}{\sum_{i=1}^{N} n_i}
\sum_{i=1}^{N}\sum_{t=1}^{n_i}
\mathrm{KL}\!\left(
p_{i,t}^{\mathcal{T}} \,\|\, p_{i,t}^{\mathcal{S}}
\right).
\label{eq:kd_loss}
\end{align}
Here $\tau>0$ is the distillation temperature. Since our corpus is in pre-training format, the loss in Eq.~\eqref{eq:kd_loss} is evaluated on all non-padding positions, rather than only on an answer segment.

Empirically, we find that optimization becomes more reliable when a small intermediate-state regularizer is added on top of token-level distillation. Following distillation practices used in the Nemotron/Minitron line~\citep{bercovich2025llamanemotron}, we introduce a low-weight cosine similarity loss on a selected set of intermediate layers $\mathcal{M}\subseteq \{1,\dots,L\}$:
\begin{align}
\mathcal{L}_{\mathrm{cos}}
=
\frac{1}{\sum_{i=1}^{N} n_i |\mathcal{M}|}
\sum_{i=1}^{N}\sum_{t=1}^{n_i}\sum_{\ell\in\mathcal{M}}
\left(
1-
\frac{
\left\langle h_{i,t}^{\mathcal{S},\ell},\, h_{i,t}^{\mathcal{T},\ell}\right\rangle
}{
\|h_{i,t}^{\mathcal{S},\ell}\|_2\,
\|h_{i,t}^{\mathcal{T},\ell}\|_2
}
\right).
\label{eq:cos_loss}
\end{align}
The final model-level objective is
\begin{align}
\mathcal{L}_{\mathrm{model}}
=
\mathcal{L}_{\mathrm{KD}}
+
\lambda_{\mathrm{cos}} \mathcal{L}_{\mathrm{cos}},
\label{eq:model_loss}
\end{align}
where $\lambda_{\mathrm{cos}}$ is a small coefficient. Intuitively, $\mathcal{L}_{\mathrm{KD}}$ restores the teacher's predictive behavior at the token level, while $\mathcal{L}_{\mathrm{cos}}$ provides a weak geometric regularization on the internal representation trajectory, which is especially helpful during the early phase of end-to-end training.

\paragraph{Discussions.}

The overall procedure can be viewed as a progressive path from \emph{intermediate-state distillation} to \emph{output-level distillation}. Stage I stabilizes optimization by solving local, feature matching problem for the edited attention blocks. Stage II then restores the student as a coherent autoregressive model under its own hidden-state dynamics. This design is particularly suitable for attention editing because the architectural perturbation is localized yet potentially severe: randomly initialized edited attention blocks are difficult to optimize from logits alone, but become trainable once they are first anchored to teacher-provided intermediate signals. Our training pipeline is inspired by that of Llama-Nemotron \cite{bercovich2025llamanemotron}. Although both approaches adopt distillation from the block level to the model level, several notable differences remain. Our objective is to introduce fundamental modifications to the attention architecture, and we retrain nearly all attention parameters from scratch, resulting in greater generality. In contrast, Llama-Nemotron prunes the existing number of heads in GQA, without making substantive structural changes.

\section{Implementation Details}
\label{sec:details}

\subsection{Instantiation with MLA and GateSWA}
\label{sec:equipment_mla_gateswa}

We instantiate our attention editing pipeline on \texttt{Qwen3-8B} and \texttt{Qwen3-30B-A3B} \citep{qwen2025qwen3}, which together provide a representative pair of open-weight hybrid-thinking backbones. This choice allows us to evaluate the proposed training procedure under both dense and mixture-of-experts settings while keeping the post-training objective unchanged. One principle in all edited variants is that we \emph{always keep the original output projection} (\texttt{o\_proj}) unchanged, which we find could perform better. To enable exact reuse of \texttt{o\_proj} (i.e., $W^O$), we require the concatenated attention output is shape-compatible with the inherited \texttt{o\_proj}. All remaining attention parameters, unless otherwise stated, are randomly initialized and subsequently optimized by the progressive distillation procedure described in Section~\ref{sec:method}.

\begin{table}[t]
\centering
\caption{Comparison (\%) of KV cache per token among GQA, MLA, and GateSWA. We follow the notations in DeepSeek-V2 \citep{liu2024deepseekv2}. Here the approximation ignores the bounded local-window ($w=128$) cache in SWA layers and only accounts for the unbounded cache carried by full-attention layers. Under a 5:1 sliding-to-full schedule, one out of every six layers is a full-attention layer.}
\label{tab:kv_cache_details}
\vspace{+0.05in}
\renewcommand{\arraystretch}{1.6}
\setlength{\tabcolsep}{16pt}
\small
\begin{tabular}{cccc}
\specialrule{1.2pt}{0pt}{0pt}
\multirow{2}{*}{\textbf{Attention Mechanisms}} & \multirow{2}{*}{KV cache per token} & \multicolumn{2}{c}{KV memory} \\
\cline{3-4}
& & Qwen3-8B & Qwen3-30B-A3B \\
\specialrule{1.2pt}{0pt}{0pt}
GQA
& $2 n_g d_h l$
& $100\%$
& $100\%$ \\
MLA
& $(d_c + d_h^R) l$
& 28\%
& 56\% \\
GateSWA & $\approx 2 n_g d_h \cdot \frac{l}{6} $
& 17\%
& 17\% \\
\specialrule{1.2pt}{0pt}{0pt}
\end{tabular}
\vspace{-0.05in}
\end{table}

\paragraph{MLA instantiation.}
We adopt a hardware-aware configuration tailored to FlashMLA kernels \citep{flashmla2025}. Let $d_c$ denote the KV compression dimension and $d_r$ the RoPE dimension used in the key branch. We set $d_c = 512$, $d_r = 64$.
This choice is motivated by practical compatibility with FlashMLA kernels, whose decode path can be viewed as an MQA-style computation with a single KV head, key width $576$, and value width $512$. We therefore obtain an efficient MQA decode path with $n_q$ query heads and one shared KV head, while retaining the expressive advantages of multi-head queries.

Unlike the original MLA parameterization, we discard the query down-projection bottleneck, and preserve the original number of query heads from Qwen3. The rationale is to avoid unnecessarily constraining query expressiveness after a substantial architectural edit. In other words, we compress K/V aggressively for decoding efficiency, but keep Q relatively unconstrained so that the edited attention can better recover the representational capacity of the teacher. In addition, to further improve inference speed, we set the non-positional key dimension to $d_k^{\mathrm{NoPE}} = 64$.
Together, these choices yield an MLA configuration that is both kernel-friendly and sufficiently expressive: KV states are aggressively compressed for decoding, the query pathway remains high-capacity, and the inherited \texttt{o\_proj} can still be reused exactly through the dimensionality constraint.

\paragraph{GateSWA instantiation.}
Most full-attention layers are replaced by sliding-window attention, and we introduce an \textit{element-wise output gate} \citep{qiu2025gatedattention} for each attention layer. Specifically, let $o_t^{\ell} \in \mathbb{R}^{d}$ denote the SDPA output of the attention branch at position $t$ and layer $\ell$, similar to notations in Section \ref{sec:problem}. We apply a learned gate
\begin{align}
g_t^{\ell} = \sigma(W_g^{\ell} h_t^{\ell-1}),
\qquad
g_t^{\ell} \in \mathbb{R}^{d},
\end{align}
and define the gated output as
\begin{align}
\widetilde{o}_t^{\ell} = g_t^{\ell} \odot o_t^{\ell},
\label{eq:gateswa_gate}
\end{align}
where $\sigma(\cdot)$ is the element-wise sigmoid function and $\odot$ denotes element-wise multiplication. This gate is intentionally lightweight, yet it introduces additional nonlinearity at the attention output and eliminates attention sink. Aside from this gating operation, all attention-related architectural settings are kept identical to the original Qwen3 backbone.

For the sliding-window component, we use a window size of $w = 128$, following the same local-attention regime adopted in GPT-OSS and MiMo-V2-Flash \citep{openai2025gptoss_modelcard, xiao2026mimov2flash}. We further interleave sliding-window and full-attention layers with a ratio of $\text{SWA} : \text{Full} = 5 : 1$. Besides that, we keep setting the first layer to full attention. Under the window size of 128, the overall KV cache is approximately reduced to one-sixth of that of the original model. The comparison of KV cache is listed in Table \ref{tab:kv_cache_details}.

\subsection{Data Construction}

We emphasize that our distillation setup differs from the common <prompt, answer> pair used in instruction distillation. All training data are in pre-training format, and our objective does not isolate an answer span. Consequently, both the block-wise objective in Eq.~\eqref{eq:block_loss} and the model-level objective in Eq.~\eqref{eq:model_loss} are defined over the full sequence, making the procedure naturally compatible with continual pre-training or architecture conversion scenarios.

\begin{wraptable}{r}{0.40\textwidth}
\vspace{-0.8\baselineskip}
\centering
\caption{Data mix in stage I.}
\label{tab:datamix_stage1}
\renewcommand{\arraystretch}{1.3}
\setlength{\tabcolsep}{10pt}
\begin{tabular}{lc}
\specialrule{1.2pt}{0pt}{0pt}
\textbf{Data source} & Percentage \\
\specialrule{1.2pt}{0pt}{0pt}
 General domain        & 40\%    \\
 Math \& code         &  35\%    \\
 Chinese corpora      & 25\%    \\
\specialrule{1.2pt}{0pt}{0pt}
\end{tabular}
\end{wraptable}
In Stages I and II, we employed entirely different dataset configurations. In Stage I, since the attention weights were randomly initialized, lower-difficulty data were more suitable; accordingly, we used a large amount of general-domain data. In Stage II, we progressively incorporated a greater proportion of complex data related to mathematics, coding, and reasoning. These data were primarily drawn from YiZhao \citep{yizhao_github} and community open-source datasets such as  \citep{dclm}, FineWeb \citep{fineweb}, Nemotron-CC \citep{nemotroncc}, MegaMath \citep{megamath}, and StarCoder2 \citep{starcoder2}. It is worth noting that, because the full pretraining corpus of Qwen3 is not publicly available \citep{qwen2025qwen3}, the data we use may differ substantially from the true distribution of its original pretraining data. This further demonstrates the robustness of our method to variations in data quality. Our training data do not overlap with the following test set.

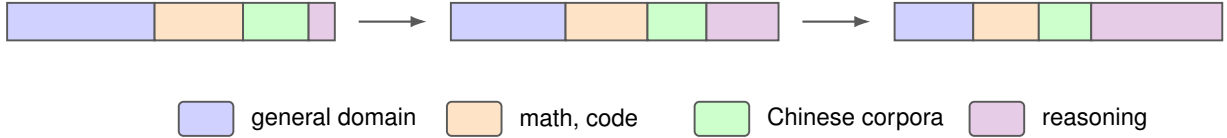
\begin{figure}[t]
    \centering
    \begin{tikzpicture}[
    font=\sffamily,
    every node/.style={font=\sffamily},
    >={Latex[length=2.4mm,width=1.7mm]},
    line width=0.8pt,
    bar/.style={draw=black!65, inner sep=0pt, outer sep=0pt},
    arrow/.style={->, draw=black!65, line width=0.9pt},
    legendbox/.style={
        draw=black!65,
        rounded corners=2pt,
        minimum width=8mm,
        minimum height=5mm,
        inner sep=0pt
    }
]

\colorlet{cgeneral}{blue!18}
\colorlet{cmath}{orange!22}
\colorlet{cmulti}{green!20}
\colorlet{ccot}{violet!20}

\def\BarL{4.8}
\def\BarH{0.55}

\coordinate (S1) at (0,0);

\pgfmathsetmacro{\sOneGen}{0.45*\BarL}
\pgfmathsetmacro{\sOneMath}{0.27*\BarL}
\pgfmathsetmacro{\sOneMulti}{0.20*\BarL}
\pgfmathsetmacro{\sOneCot}{0.08*\BarL}

\filldraw[bar, fill=cgeneral]
    (S1) rectangle ++(\sOneGen,\BarH);
\filldraw[bar, fill=cmath]
    ($(S1)+(\sOneGen,0)$) rectangle ++(\sOneMath,\BarH);
\filldraw[bar, fill=cmulti]
    ($(S1)+(\sOneGen+\sOneMath,0)$) rectangle ++(\sOneMulti,\BarH);
\filldraw[bar, fill=ccot]
    ($(S1)+(\sOneGen+\sOneMath+\sOneMulti,0)$) rectangle ++(\sOneCot,\BarH);

\coordinate (S2) at (6.5,0);

\pgfmathsetmacro{\sTwoGen}{0.35*\BarL}
\pgfmathsetmacro{\sTwoMath}{0.25*\BarL}
\pgfmathsetmacro{\sTwoMulti}{0.18*\BarL}
\pgfmathsetmacro{\sTwoCot}{0.22*\BarL}

\filldraw[bar, fill=cgeneral]
    (S2) rectangle ++(\sTwoGen,\BarH);
\filldraw[bar, fill=cmath]
    ($(S2)+(\sTwoGen,0)$) rectangle ++(\sTwoMath,\BarH);
\filldraw[bar, fill=cmulti]
    ($(S2)+(\sTwoGen+\sTwoMath,0)$) rectangle ++(\sTwoMulti,\BarH);
\filldraw[bar, fill=ccot]
    ($(S2)+(\sTwoGen+\sTwoMath+\sTwoMulti,0)$) rectangle ++(\sTwoCot,\BarH);

\coordinate (S3) at (13.0,0);

\pgfmathsetmacro{\sThreeGen}{0.24*\BarL}
\pgfmathsetmacro{\sThreeMath}{0.20*\BarL}
\pgfmathsetmacro{\sThreeMulti}{0.16*\BarL}
\pgfmathsetmacro{\sThreeCot}{0.40*\BarL}

\filldraw[bar, fill=cgeneral]
    (S3) rectangle ++(\sThreeGen,\BarH);
\filldraw[bar, fill=cmath]
    ($(S3)+(\sThreeGen,0)$) rectangle ++(\sThreeMath,\BarH);
\filldraw[bar, fill=cmulti]
    ($(S3)+(\sThreeGen+\sThreeMath,0)$) rectangle ++(\sThreeMulti,\BarH);
\filldraw[bar, fill=ccot]
    ($(S3)+(\sThreeGen+\sThreeMath+\sThreeMulti,0)$) rectangle ++(\sThreeCot,\BarH);

\draw[arrow] ($(S1)+(\BarL+0.35,0.5*\BarH)$) -- ($(S2)+(-0.35,0.5*\BarH)$);
\draw[arrow] ($(S2)+(\BarL+0.35,0.5*\BarH)$) -- ($(S3)+(-0.35,0.5*\BarH)$);

\coordinate (L0) at (2.5,-1.15);

\node[legendbox, fill=cgeneral, anchor=west] (lg1) at (L0) {};
\node[anchor=west] at ($(lg1.east)+(0.12,0)$) {general domain};

\node[legendbox, fill=cmath, anchor=west] (lg2) at ($(lg1.east)+(3.1,0)$) {};
\node[anchor=west] at ($(lg2.east)+(0.12,0)$) {math, code};

\node[legendbox, fill=cmulti, anchor=west] (lg3) at ($(lg2.east)+(2.8,0)$) {};
\node[anchor=west] at ($(lg3.east)+(0.12,0)$) {Chinese corpora};

\node[legendbox, fill=ccot, anchor=west] (lg4) at ($(lg3.east)+(3.2,0)$) {};
\node[anchor=west] at ($(lg4.east)+(0.12,0)$) {reasoning};

\end{tikzpicture}
    \vspace{+0.05in}
    \caption{Data mix in stage II. At this stage, curriculum learning was employed to control the data composition, gradually increasing the proportion of reasoning-related data while reducing that of general domains. During training, the optimizer state keeps preserved across changes in the data mixture.}
    \label{fig:datasets_illustration}
\end{figure}

In Stage I, the vast majority of the data consisted of general-domain content, and the overall proportions are shown in Table \ref{tab:datamix_stage1}. Our data categories mainly include general-domain, mathematics and code, and Chinese corpora. These relatively simple data were used to enable the model to optimize smoothly from random initialization. At this stage, we used 2B tokens and trained with a sequence length of 8K.

In Stage II, the dataset still comprised the same four categories as in Stage I. In our experiments, we found that if we directly used a larger proportion of more difficult corpora, such as complex mathematics, code, and reasoning data, the model was often difficult to optimize. Conversely, using a larger amount of easier data leads to inferior model performance. We therefore adopt a curriculum learning strategy, in which three groups of data with progressively increasing difficulty are concatenated, as illustrated in Figure \ref{fig:datasets_illustration}, with the proportion of more challenging data gradually increasing. During training, when switching between datasets, the optimizer states are preserved. In this stage, we use a total of 6B tokens, arranged as three consecutive segments of 2B tokens each, as illustrated in Figure \ref{fig:datasets_illustration}. It is noteworthy that the amount of training data used in our study is less than one-thousandth of that used for Qwen3.


\section{Experimental Results}
\label{sec:experiments}

In this section, we present the experimental results, focusing on factors such as model performance, inference speed, and improvements in hardware utilization efficiency. Since the weights of the modified attention modules are randomly initialized, we first verify that the proposed method can recover the base-model performance, mainly through a set of few-shot evaluations of fundamental capabilities. In addition, because our target model is a hybrid reasoning model, we also report several chat- and reasoning-related benchmarks to assess its effectiveness. Finally, we provide a detailed analysis of the inference acceleration achieved by the modified architecture.

\subsection{Model Performance}

\begin{table}[b]
\centering
\vspace{-0.10in}
\caption{Performance (\%) of different attention variants under few-shot evaluation. The model is required to directly output the answer, and accuracy is computed by matching the predicted answer against the ground truth.
}
\vspace{+0.05in}
\label{tab:model_few_shot_results}
\renewcommand{\arraystretch}{1.5}
\setlength{\tabcolsep}{18pt} 
\begin{tabular}{lccccccc}
\specialrule{1.2pt}{0pt}{0pt}
\textbf{Models} & ARC-E & ARC-C & C-Eval & MMLU \\
\specialrule{1.2pt}{0pt}{0pt}
Qwen3-8B            & 96.46 & 91.46 & 77.04 & 74.47  \\
Qwen3-8B-GateSWA    & 96.54 & 90.35 & 76.82 & 73.26 \\
Qwen3-8B-MLA        & 96.71 & 90.01 & 73.70 & 71.77  \\
\midrule 
Qwen3-30B-A3B          & 96.89 & 92.58 & 83.06  & 78.68 \\
Qwen3-30B-A3B-GateSWA  & 98.19 & 91.98 & 81.20  & 77.24\\
Qwen3-30B-A3B-MLA      & 98.15 & 93.77 & 81.05  & 76.99 \\
\specialrule{1.2pt}{0pt}{0pt}
\end{tabular}
\end{table}

\paragraph{Pre-training evaluation.}
To validate that the attention editing has recovered the basic capacity, we first test the few-shot performance. We conduct evaluation in a few-shot setting, requiring the model to directly output one token as the answer (e.g., one of ``A'', ``B'', ``C'', or ``D''), which is then extracted and matched against the ground truth \citep{zheng2024sglang}. Since nearly all weights in the attention modules are randomly initialized, the model is initially unable to answer any questions. This evaluation therefore serves to verify whether the model parameters have been effectively optimized, particularly with respect to its memory-related capabilities. Concretely, the few-shot performance benchmarks include ARC-asy, ARC-challenge \citep{arc}, MMLU \citep{mmlu}, and C-Eval \citep{ceval}. The results are listed in Table \ref{tab:model_few_shot_results}.

 As shown in Table \ref{tab:model_few_shot_results}, the fundamental capabilities are well preserved, particularly on relatively simple memory-related benchmarks. This demonstrates the effectiveness of the progressive distillation method proposed in this paper, indicating that it can indeed transform one attention architecture into another. For example, on ARC-Easy, Qwen3-8B-MLA outperforms the original Qwen3-8B by 0.25\%. This is a rather interesting phenomenon, and we speculate that the following explanation may account for it: prior works \citep{geva2021ffnmemory,dai2022knowledgeneurons,meng2022rome,niu2024knthesis} suggest that the attention module primarily serves as a mechanism for knowledge retrieval, whereas the FFN module is regarded as the locus of knowledge storage. Since we leave all FFN parameters frozen, the model’s knowledge storage remains intact, and only the retrieval functionality needs to be relearned. Gated attention is often considered to enhance nonlinearity and mitigate the attention sink phenomenon \citep{qiu2025gatedattention}, which may in some cases improve retrieval capability. MLA, in contrast, is generally regarded as having stronger expressive capacity than GQA, and may therefore also yield performance gains in certain settings \citep{liu2024deepseekv2,deepseekv3}.

\paragraph{Post-training evaluation.}

In this part, we evaluate the model’s chat and reasoning capabilities, noting that these are abilities not possessed by the previously converted models. We aim to examine whether the method proposed in this paper can recover, to some extent, the model’s capacity for chat and even reasoning. Specifically, we primarily adopt GSM8K \citep{gsm8k}, a mathematics benchmark of moderate difficulty, and C-Eval \citep{ceval}, a more diverse general-domain benchmark covering a broad range of subjects. In addition, we report results under both thinking-enabled modes.


%

\begin{wraptable}{r}{0.52\textwidth}
\vspace{-0.8\baselineskip}
\centering
\caption{Accuracy (\%) of different attention-variant models on GSM8K and C-Eval in thinking-mode.}
\label{tab:result_chat_reasoning}
\renewcommand{\arraystretch}{1.5}
\setlength{\tabcolsep}{6pt}
\begin{tabular}{lcc}
\specialrule{1.2pt}{0pt}{0pt}
\textbf{Models} & \makecell[c]{\rule{0pt}{2.6ex}GSM8K\\(thinking)} & \makecell[c]{\rule{0pt}{2.6ex}C-Eval\\(thinking)} \\
\specialrule{1.2pt}{0pt}{0pt}
 Qwen3-8B           &  95.98   &  83.80   \\
 Qwen3-8B-GateSWA   &  94.31   &  78.90        \\
 Qwen3-8B-MLA       &  95.00   &  79.20   \\
 \hline
 Qwen3-30B               & 96.44   & 86.40   \\
 Qwen3-30B-A3B-GateSWA   & 95.15   & 81.28   \\
 Qwen3-30B-A3B-MLA       & 94.39   & 81.13    \\
\specialrule{1.2pt}{0pt}{0pt}
\end{tabular}
\end{wraptable}
The results are listed in Table \ref{tab:result_chat_reasoning}, and we adopt AISBench\footnote{\url{https://github.com/AISBench/benchmark}} tool to conduct evaluation. To our surprise, on the GSM8K task, the model with the modified attention architecture is able to achieve an accuracy remarkably close to that of the original 8B model. \textit{Qwen3 series underwent 36 trillion tokens of pre-training, along with highly sophisticated post-training}, whereas our method attains comparable performance using \textit{fewer than 10B tokens}. Moreover, the training data we use are expected to differ substantially in distribution from those used for Qwen3. Although GSM8K is not considered as a challenging benchmark by current standards, these results nevertheless provide clear evidence for the success of our attention editing approach. On a general knowledge benchmark such as C-Eval, the converted model exhibits a certain performance gap. We attribute this to the fact that knowledge-oriented tasks depend more directly on the model’s internal knowledge, whereas mathematical tasks are more amenable to compensation through chain-of-thought reasoning.

\subsection{Inference Speedup}
\label{sec:inference_acceleration}

In the previous part, we showed that progressive distillation enables the converted models to recover strong general-domain and reasoning capabilities. We now turn to evaluate whether attention conversion also delivers the intended \emph{hardware-level inference benefits}. Our goal in this part is to verify that, beyond maintaining model performance, the edited attention architectures substantially improve deployment efficiency.

\begin{figure*}[t]
    \centering
    \includegraphics[width=0.98\textwidth]{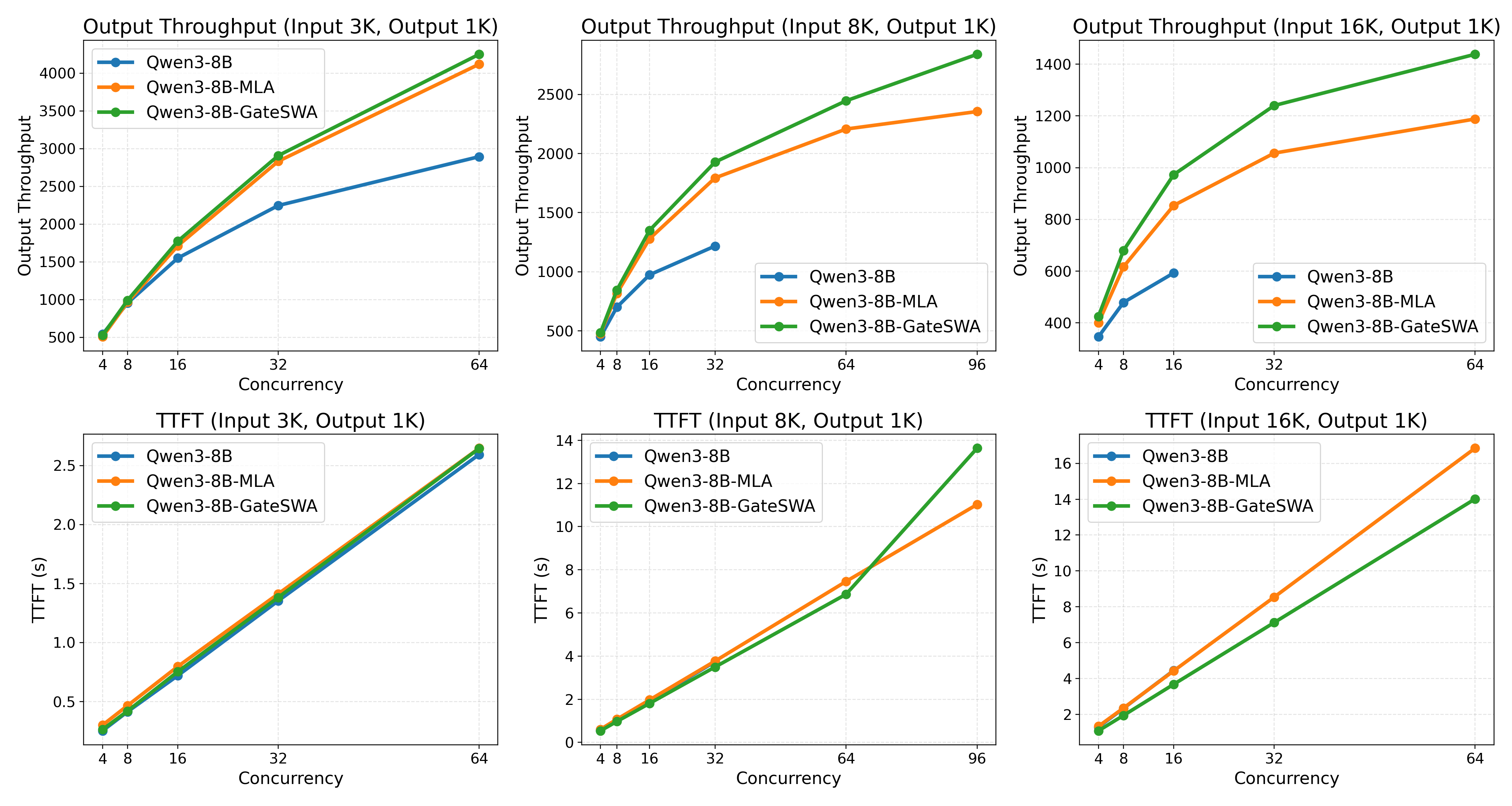}
    \caption{Performance under different concurrency levels for three input lengths. The \textit{upper row} shows the variation in throughput , while the \textit{lower row} shows the variation in TTFT. The experimental setup is as follows: inference is conducted on a single \textit{\textbf{H800}} GPU using vLLM version 0.16.0, with gpu-memory-utilization set to 0.8. Missing data in the figure indicate that the model is unable to handle that level of concurrency simultaneously.}
    \label{fig:throughput-ttft-all}
\end{figure*}

\begin{figure*}[t]
    \centering
    \includegraphics[width=0.90\textwidth]{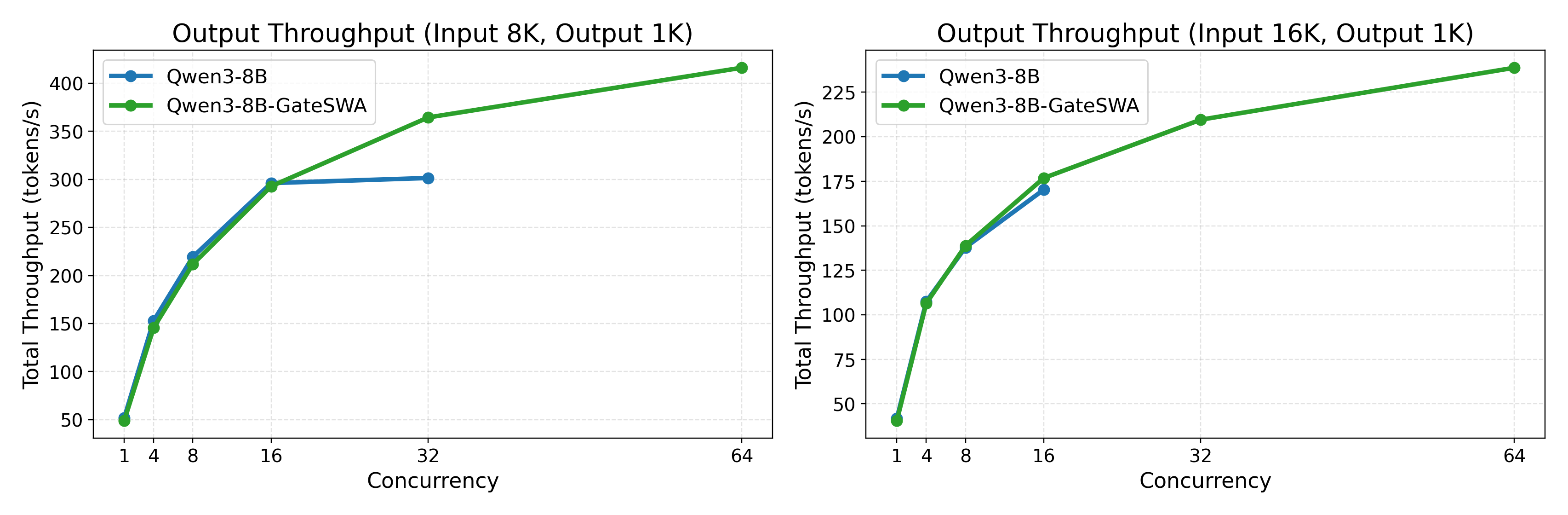}
    \caption{Throughput under different concurrency levels. The experimental setup is as follows: inference is conducted on a single \textit{\textbf{Ascend 910B}} using vLLM/vLLM-Ascend version 0.16.0. Missing data in the figure indicate that the model is unable to handle that level of concurrency simultaneously. The results about Qwen3-8B-MLA are missing, because the MLA kernels in higher-version vLLM-Ascend impose customized requirements on specific parameter configuration.}
    \label{fig:throughput-910B}
    \vspace{-0.05in}
\end{figure*}

\paragraph{Evaluation protocol.}
We evaluate inference performance under a controlled decoding setup and report two standard serving metrics: \emph{time to first token} (TTFT) and \emph{output throughput}. TTFT measures the elapsed time between the arrival of a request and the emission of its first output token, and therefore captures the latency perceived by the user before generation starts. Output throughput measures the total number of generated output tokens per unit time across all served requests, and reflects the steady-state decoding efficiency of the serving system.

These two metrics are complementary: TTFT emphasizes responsiveness, whereas output throughput emphasizes sustained generation capacity under load. We benchmark the original and converted models under three input lengths, namely 3K/8K/16K tokens, while fixing the output length to 1K tokens. For each input length, we vary the request concurrency and record both TTFT and output throughput.

\paragraph{KV-cache reduction.}
The memory benefit of attention conversion is summarized in Table~\ref{tab:kv_cache_details}. For example, the converted models Qwen3-8B-GateSWA reduce KV-cache usage by approximately $80\%$ relative to the original full-attention backbone, which is a substantial reduction. From a systems perspective, such a drop in KV-cache footprint directly increases the effective memory budget available for active requests and long contexts. In practice, this means that the same hardware can either accommodate more cached tokens per request or serve more concurrent requests before hitting memory limits, both of which translate into lower serving cost and improved deployment flexibility.

\paragraph{Decoding throughput and TTFT.}
The throughput and TTFT results are shown in Figure~\ref{fig:throughput-ttft-all}-\ref{fig:throughput-910B}. The advantage of inference speed becomes increasingly pronounced as concurrency grows. This trend is consistent with the intended systems effect of attention conversion. When concurrency is low, the GPU is less constrained by KV-cache residency and memory traffic, so the benefit of reducing the attention-state footprint is present but limited. As concurrency increases, however, the decoding system must maintain more active sequences simultaneously, and KV-cache pressure becomes increasingly dominant. In this regime, the converted models enjoy a large advantage because they require substantially less memory per request and incur lower cache-management overhead.

For TTFT, performance depends more heavily on input tokens processed during the prefill stage and is typically compute-bound. In the TTFT plot, the trends of MLA and the original model largely overlap. For MLA, the prefill stage follows an MHA-style computation, and although its FLOPs are lower than those of an MHA model of comparable size, they are not necessarily lower than those of GQA. Consequently, MLA does not exhibit a clear advantage in TTFT. By contrast, GateSWA substantially reduces computational cost through the sliding-window mechanism, while the additional computation introduced by the gating operation is relatively minor, leading to a more noticeable improvement in TTFT, which is consistent with the tendency shown in Figure \ref{fig:throughput-ttft-all}.

These results show that the proposed attention conversion could preserve model capability after distillation, and it also yields a meaningful systems-level payoff. The converted models substantially reduce KV-cache consumption and, as a consequence, deliver higher decoding throughput and better TTFT. The gain is especially evident when concurrency is high, which is precisely the operating regime where deployment cost and serving efficiency matter most.

\section{Conclusion and Future Work}
\label{sec:conclusions}

In this paper, we present Attention Editing, a general framework for converting the attention architecture of a trained LLM without re-pretraining from scratch. By treating the target attention as a learnable replacement, and training it with progressive distillation, our method avoids delicate architecture-specific weight surgery and enables substantial post hoc attention refactoring. Experiments on converting GQA-based Qwen3 models to both MLA and GateSWA show that competitive quality can be retained while improving efficiency. In addition, all experiments were conducted entirely on an Ascend 910B cluster, providing a practical case study of large-model post-training on domestic hardware.

Although some current results are promising, several important questions remain open. First, the amount of training data used in our study is less than \textit{one-thousandth of that used for Qwen3}, so it remains unclear how Attention Editing will scale with substantially larger and more diverse data. Understanding whether further scaling can yield stronger recovery and better downstream efficiency-quality trade-offs is an especially interesting direction. Second, our current training mixture contains relatively limited a\textit{gent-oriented data}, and our evaluation focuses more on chat and reasoning abilities than on fully interactive agent performance. Extending attention editing toward stronger tool-use and long-horizon agent capabilities is therefore a natural next step.

\bibliographystyle{unsrtnat}
\bibliography{references} 

@inproceedings{agarwal2024gkd,
  title     = {On-Policy Distillation of Language Models: Learning from Self-Generated Mistakes},
  author    = {Agarwal, Rishabh and Vieillard, Nino and Zhou, Yongchao and Stanczyk, Piotr and Ramos, Sabela and Geist, Matthieu and Bachem, Olivier},
  booktitle = {International Conference on Learning Representations},
  year      = {2024},
}

@article{ainslie2023gqa,
  title   = {GQA: Training Generalized Multi-Query Transformer Models from Multi-Head Checkpoints},
  author  = {Ainslie, Joshua and Lee-Thorp, James and de Jong, Michiel and Zemlyanskiy, Yury and Lebr{\'o}n, Federico and Sanghai, Sumit},
  journal = {arXiv preprint arXiv:2305.13245},
  year    = {2023},
}

@article{gu2023minillm,
  title   = {MiniLLM: Knowledge Distillation of Large Language Models},
  author  = {Gu, Yuxian and others},
  journal = {arXiv preprint arXiv:2306.08543},
  year    = {2023},
}

@article{hinton2015distill,
  title   = {Distilling the Knowledge in a Neural Network},
  author  = {Hinton, Geoffrey and Vinyals, Oriol and Dean, Jeff},
  journal = {arXiv preprint arXiv:1503.02531},
  year    = {2015},
}

@article{ji2025mha2mla,
  title   = {Towards Economical Inference: Enabling DeepSeek's Multi-Head Latent Attention in Any Transformer-based LLMs},
  author  = {Ji, Tao and Guo, Bin and Wu, Yuanbin and Guo, Qipeng and Shen, Lixing and Chen, Zhan and Qiu, Xipeng and Zhang, Qi and Gui, Tao},
  journal = {arXiv preprint arXiv:2502.14837},
  year    = {2025},
}

@inproceedings{katharopoulos2020linear,
  title     = {Transformers are RNNs: Fast Autoregressive Transformers with Linear Attention},
  author    = {Katharopoulos, Angelos and Vyas, Apoorv and Pappas, Nikolaos and Fleuret, Fran{\c{c}}ois},
  booktitle = {International Conference on Machine Learning},
  year      = {2020},
}

@techreport{koike2026latentllm,
  title       = {LatentLLM: Activation-Aware Transform to Multi-Head Latent Attention},
  author      = {Koike-Akino, Toshiaki and Chen, Xiangyu and Liu, Jing and Wang, Ye and Wang, Pu and Brand, Matthew},
  institution = {Mitsubishi Electric Research Laboratories},
  year        = {2026},
}

@article{kwon2023pagedattention,
  title   = {Efficient Memory Management for Large Language Model Serving with PagedAttention},
  author  = {Kwon, Woosuk and Li, Zhuohan and Zhuang, Siyuan and Sheng, Ying and Zheng, Lianmin and Yu, Cody Hao and Gonzalez, Joseph E and Zhang, Hao and Stoica, Ion},
  journal = {arXiv preprint arXiv:2309.06180},
  year    = {2023},
}

@misc{liu2024deepseekv2,
      title={DeepSeek-V2: A Strong, Economical, and Efficient Mixture-of-Experts Language Model}, 
      author={DeepSeek-AI},
      year={2024},
      eprint={2405.04434},
      archivePrefix={arXiv},
      primaryClass={cs.CL}
}

@article{meng2025transmla,
  title   = {TransMLA: Multi-Head Latent Attention Is All You Need},
  author  = {Meng, Fanxu and Tang, Pingzhi and Tang, Xiaojuan and Yao, Zengwei and Sun, Xing and Zhang, Muhan},
  journal = {arXiv preprint arXiv:2502.07864},
  year    = {2025},
}

@article{romero2014fitnets,
  title   = {FitNets: Hints for Thin Deep Nets},
  author  = {Romero, Adriana and Ballas, Nicolas and Ebrahimi Kahou, Samira and Chassang, Antoine and Gatta, Carlo and Bengio, Yoshua},
  journal = {arXiv preprint arXiv:1412.6550},
  year    = {2014},
}

@article{shazeer2019fast,
  title   = {Fast Transformer Decoding: One Write-Head is All You Need},
  author  = {Shazeer, Noam},
  journal = {arXiv preprint arXiv:1911.02150},
  year    = {2019},
}

@inproceedings{vaswani2017attention,
  title     = {Attention Is All You Need},
  author    = {Vaswani, Ashish and Shazeer, Noam and Parmar, Niki and Uszkoreit, Jakob and Jones, Llion and Gomez, Aidan N and Kaiser, Lukasz and Polosukhin, Illia},
  booktitle = {Advances in Neural Information Processing Systems},
  year      = {2017},
}

@misc{openai2025gptoss_modelcard,
  title        = {gpt-oss-120b \& gpt-oss-20b Model Card},
  author       = {{OpenAI}},
  year         = {2025},
  howpublished = {Model card},
}

@article{yao2022react,
  title   = {ReAct: Synergizing Reasoning and Acting in Language Models},
  author  = {Yao, Shunyu and Zhao, Jeffrey and Yu, Dian and Du, Nan and Shafran, Izhak and Narasimhan, Karthik and Cao, Yuan},
  journal = {arXiv preprint arXiv:2210.03629},
  year    = {2022},
}

@article{schick2023toolformer,
  title   = {Toolformer: Language Models Can Teach Themselves to Use Tools},
  author  = {Schick, Timo and Dwivedi-Yu, Jane and Dess{\`i}, Roberto and Raileanu, Roberta and Lomeli, Maria and Zettlemoyer, Luke and Cancedda, Nicola and Scialom, Thomas},
  journal = {arXiv preprint arXiv:2302.04761},
  year    = {2023},
}

@article{beltagy2020longformer,
  title   = {Longformer: The Long-Document Transformer},
  author  = {Beltagy, Iz and Peters, Matthew E. and Cohan, Arman},
  journal = {arXiv preprint arXiv:2004.05150},
  year    = {2020},
}

@inproceedings{choromanski2021performer,
  title     = {Rethinking Attention with Performers},
  author    = {Choromanski, Krzysztof and Likhosherstov, Valerii and Dohan, David and Song, Xingyou and Gane, Andreea and Sarlos, Tamas and Hawkins, Peter and Davis, Jared and Mohiuddin, Afroz and Kaiser, Lukasz and Belanger, David and Colwell, Lucy and Weller, Adrian},
  booktitle = {International Conference on Learning Representations},
  year      = {2021},
}

@article{xiao2023streamingllm,
  title   = {Efficient Streaming Language Models with Attention Sinks},
  author  = {Xiao, Guang and others},
  journal = {arXiv preprint arXiv:2309.17453},
  year    = {2023},
}

@article{qiu2025gatedattention,
  title   = {Gated Attention for Large Language Models: Non-linearity, Sparsity, and Attention-Sink-Free},
  author  = {Qiu, Zihan and Wang, Zekun and Zheng, Bo and Huang, Zeyu and Wen, Kaiyue and Yang, Songlin and Men, Rui and Yu, Le and Huang, Fei and Huang, Suozhi and Liu, Dayiheng and Zhou, Jingren and Lin, Junyang},
  journal = {arXiv preprint arXiv:2505.06708},
  year    = {2025},
}

@article{xiao2026mimov2flash,
  title   = {MiMo-V2-Flash Technical Report},
  author  = {{LLM-Core Xiaomi}},
  journal = {arXiv preprint arXiv:2601.02780},
  year    = {2026},
}

@article{qwen2025qwen3,
    title={Qwen3 Technical Report}, 
    author={Yang, An and others},
    journal = {arXiv preprint arXiv:2505.09388},
    year={2025}
}

@inproceedings{ouyang2022instructgpt,
  title     = {Training language models to follow instructions with human feedback},
  author    = {Ouyang, Long and Wu, Jeff and Jiang, Xu and Almeida, Diogo and Wainwright, Carroll L. and Mishkin, Pamela and Zhang, Chong and Agarwal, Sandhini and Slama, Katarina and Ray, Alex and Schulman, John and Hilton, Jacob and Leike, Jan and Lowe, Ryan},
  booktitle = {Advances in Neural Information Processing Systems},
  year      = {2022},
}

@article{deepseekv3,
      title={DeepSeek-V3 Technical Report}, 
      author={DeepSeek-AI},
      year={2024},
      journal={arXiv preprint arXiv:2412.19437},
}

@article{deepseekr1,
      title={DeepSeek-R1: Incentivizing Reasoning Capability in LLMs via Reinforcement Learning}, 
      author={DeepSeek-AI},
      year={2025},
      journal={arXiv preprint arXiv:2501.12948},
}

@inproceedings{yang2025gateddeltanet,
  title     = {Gated Delta Networks: Improving Mamba2 with Delta Rule},
  author    = {Yang, Songlin and Kautz, Jan and Hatamizadeh, Ali},
  booktitle = {International Conference on Learning Representations},
  year      = {2025},
}

@article{zhang2025kimilinear,
  title         = {Kimi Linear: An Expressive, Efficient Attention Architecture},
  author        = {Zhang, Yu and Lin, Zongyu and Yao, Xingcheng and others},
  year          = {2025},
  journal        = {arXiv preprint arXiv:2510.26692},
}

@inproceedings{bercovich2025puzzle,
  title     = {Puzzle: Distillation-Based NAS for Inference-Optimized LLMs},
  author = {Bercovich, Akhiad and Ronen, Tomer and Abramovich, Talor and others},
  booktitle = {International Conference on Machine Learning},
  year      = {2025},
}

@article{bercovich2025llamanemotron,
  title         = {Llama-Nemotron: Efficient Reasoning Models},
  author        = {Bercovich, Akhiad and Levy, Itay and Golan, Izik and others},
  year          = {2025},
  journal        = {arXiv preprint arXiv:2505.00949},
}

@article{gu2025attentionsink,
  title   = {When Attention Sink Emerges in Language Models: An Empirical View},
  author  = {Gu, Xiangming and Pang, Tianyu and Du, Chao and Liu, Qian and Zhang, Fengzhuo and Du, Cunxiao and Wang, Ye and Lin, Min},
  journal = {arXiv preprint arXiv:2410.10781},
  year    = {2025},
}

@misc{flashmla2025,
  title        = {FlashMLA: Efficient Multi-head Latent Attention Kernels},
  author       = {Jiashi Li and Shengyu Liu},
  year         = {2025},
  publisher    = {GitHub},
}

@misc{yizhao_github,
  author       = {{HITsz-TMG}},
  title        = {YiZhao: A 2TB Open Financial Corpus},
  publisher    = {GitHub},
  year      = {2026}
}

@inproceedings{zheng2024sglang,
  title     = {SGLang: Efficient Execution of Structured Language Model Programs},
  author    = {Zheng, Lianmin and others},
  booktitle = {Advances in Neural Information Processing Systems},
  year      = {2024}
}

@misc{qwen3next80ba3b,
  title        = {Qwen3-Next-80B-A3B-Instruct},
  author       = {Qwen Team},
  year         = {2025},
  publisher    = {Hugging Face},
}

@article{kimiteam2025kimik2openagentic,
  title         = {Kimi K2: Open Agentic Intelligence},
  author        = {Kimi Team},
  year          = {2025},
  journal        = {arXiv preprint arXiv:2507.20534},
}

@article{glm5team2026glm5,
  title   = {GLM-5: from Vibe Coding to Agentic Engineering},
  author  = {GLM-5 Team},
  journal = {arXiv preprint arXiv:2602.15763},
  year    = {2026},
}

@article{peng2023rwkv,
  title   = {RWKV: Reinventing RNNs for the Transformer Era},
  author  = {Peng, Bo and others},
  journal = {arXiv preprint arXiv:2305.13048},
  year    = {2023}
}

@article{mamba,
  title   = {Mamba: Linear-Time Sequence Modeling with Selective State Spaces},
  author  = {Gu, Albert and Dao, Tri},
  journal = {arXiv preprint arXiv:2312.00752},
  year    = {2023}
}

@article{stepfun2026step35flash,
  title   = {Step 3.5 Flash: Open Frontier-Level Intelligence with 11B Active Parameters},
  author  = {StepFun Team},
  journal = {arXiv preprint arXiv:2602.10604},
  year    = {2026}
}

@article{starcoder2,
  title   = {StarCoder2 and The Stack v2: The Next Generation},
  author  = {Lozhkov, Anton and others},
  journal = {arXiv preprint arXiv:2402.19173},
  year    = {2024}
}

@article{dclm,
  title   = {DataComp-LM: In Search of the Next Generation of Training Sets for Language Models},
  author  = {Li, Jeffrey and others},
  journal = {arXiv preprint arXiv:2406.11794},
  year    = {2024}
}

@article{fineweb,
  title   = {Decanting the Web for the Finest Text Data at Scale},
  author  = {Penedo, Guilherme and others},
  journal = {arXiv preprint arXiv:2406.17557},
  year    = {2024}
}

@article{megamath,
  title   = {MegaMath: Pushing the Limits of Open Math Corpora},
  author  = {Zhou, Fan and others},
  journal = {arXiv preprint arXiv:2504.02807},
  year    = {2025}
}

@article{nemotroncc,
  title   = {Nemotron-CC: Transforming Common Crawl into a Refined Long-Horizon Pretraining Dataset},
  author  = {Su, Dan and others},
  journal = {arXiv preprint arXiv:2412.02595},
  year    = {2024}
}

@article{arc,
  title   = {Think you have Solved Question Answering? Try ARC, the AI2 Reasoning Challenge},
  author  = {Clark, Peter and others},
  journal = {arXiv preprint arXiv:1803.05457},
  year    = {2018}
}

@article{ceval,
  title   = {C-Eval: A Multi-Level Multi-Discipline Chinese Evaluation Suite for Foundation Models},
  author  = {Huang, Yuzhen and others},
  journal = {arXiv preprint arXiv:2305.08322},
  year    = {2023}
}

@inproceedings{mmlu,
  title     = {Measuring Massive Multitask Language Understanding},
  author    = {Hendrycks, Dan and others},
  booktitle = {International Conference on Learning Representations},
  year      = {2021}
}

@article{gsm8k,
  title   = {Training Verifiers to Solve Math Word Problems},
  author  = {Cobbe, Karl and others},
  journal = {arXiv preprint arXiv:2110.14168},
  year    = {2021}
}

@inproceedings{geva2021ffnmemory,
	title     = {Transformer Feed-Forward Layers Are Key-Value Memories},
	author    = {Geva, Mor and others},
	booktitle = {Proceedings of the Conference on Empirical Methods in Natural Language Processing},
	year      = {2021}
}

@inproceedings{dai2022knowledgeneurons,
	title     = {Knowledge Neurons in Pretrained Transformers},
	author    = {Dai, Damai and others},
	booktitle = {Proceedings of the Annual Meeting of the Association for Computational Linguistics},
	year      = {2022}
}

@inproceedings{meng2022rome,
	title     = {Locating and Editing Factual Associations in GPT},
	author    = {Meng, Kevin and others},
	booktitle = {Advances in Neural Information Processing Systems},
	year      = {2022}
}

@inproceedings{niu2024knthesis,
	title     = {What does the Knowledge Neuron Thesis Have to do with Knowledge?},
	author    = {Niu, Jingcheng and others},
	booktitle = {International Conference on Learning Representations},
	year      = {2024}
}

\end{document}